\documentclass[11pt]{article}

\usepackage[final]{acl}

\usepackage{times}
\usepackage{latexsym}
\usepackage{blindtext}
\usepackage[T1]{fontenc}
\usepackage{booktabs}
\usepackage{xcolor}
\usepackage{listings}
\usepackage{amsmath}
\usepackage{longtable}
\usepackage{array}
\lstdefinestyle{promptstyle}{
    basicstyle=\ttfamily\footnotesize,
    breaklines=true,
    breakatwhitespace=true,
    columns=fullflexible,
    keepspaces=true,
    frame=single,
    framerule=0.3pt,
    rulecolor=\color{gray},
    backgroundcolor=\color{gray!4},
    xleftmargin=0pt,
    xrightmargin=0pt,
    aboveskip=6pt,
    belowskip=6pt,
    literate=
      {→}{{$\rightarrow$}}1
      {—}{{--}}1
      {–}{{--}}1
      {“}{{}}1
      {”}{{}}1
      {’}{{'}}1
}
\usepackage[utf8]{inputenc}

\usepackage{microtype}

\usepackage{inconsolata}

\usepackage{graphicx}

\title{Mapping the Afterlife of Future Work Statements in NLP Research}

\author{Ioana-Roxana Boriceanu, Claudiu Creanga, Ana Sabina Uban \and Liviu P Dinu \\
  Faculty of Mathematics and Informatics, University of Bucharest \\
  \texttt{ioana-roxana.boriceanu@s.unibuc.ro, claudiu.creanga@fmi.unibuc.ro} \\
  \texttt{auban@fmi.unibuc.ro, ldinu@fmi.unibuc.ro}}

\begin{document}
\maketitle

%----------------------------------------------------------------------------------%
\begin{abstract}
Future work statements are one of the few places in scientific papers where authors explicitly describe what they think should happen next. We study what happens to such statements after publication. We construct a dataset of 72,070 future work items extracted from 23,113 ACL Anthology papers published between 1979 and 2025, categorize them by type, retrieve candidate follow-up papers through citation-based and content-based search, and use LLM-based judgments to assess whether later papers pursue the proposed directions. Using this resource, we analyze patterns across time, venues, categories, themes, and authors. We find that many items are later addressed, often within a few years, with clear differences across types and themes and a largely community-driven pattern. Our results suggest that future work statements provide useful evidence for studying how research directions develop in NLP. To support further research, we release our dataset at: https://doi.org/10.5281/zenodo.20385239.
\end{abstract}

%----------------------------------------------------------------------------------%
\section{Introduction}
Scientific papers are usually read as records of completed work: a problem is introduced, a method is proposed, and evidence is presented. Yet many papers also point beyond the work they report. In future work statements, authors identify limitations, propose extensions, suggest applications, and name problems that remain open. These statements are often brief, but they are one of the few places where researchers explicitly state what they think should happen next. This makes future work statements useful for studying how research agendas develop. A proposed direction may be pursued quickly, taken up years later, addressed by other researchers, or remain unaddressed. The central question is therefore not only what future work statements say, but what happens to them after publication.

Existing work gives us partial tools for answering this question. Scientific discourse mining has shown that papers contain structured signals such as citation intent, rhetorical roles, limitations, and future work statements \citep{cohan2019structural,hu2015mining,zhang2023automatic}. Work on scientific change has traced fields through citations, topics, rhetorical framing, and relations between ideas \citep{hall2008studying,prabhakaran2016predicting,jurgens2018measuring,tan2017friendships}. These approaches show that scientific text can reveal both local discourse functions and broader research dynamics. What they do not directly capture is whether explicit future work statements are later addressed by subsequent research.

Natural language processing (NLP) is a useful setting for this study because it has a long shared publication record in the ACL Anthology and changes rapidly across tasks, methods, and evaluation practices. Future work statements in this literature range from concrete next steps to harder open problems. Tracking them over time lets us study which stated directions later appear in follow-up research.

In this paper, we treat future work statements as explicit signals of research directions. We construct a dataset of 72,070 future work items extracted from 23,113 ACL Anthology papers published between 1979 and 2025. Each item is categorized by type and linked to candidate follow-up papers retrieved through citation-based and content-based search. We then determine whether later papers address each item and analyze patterns across time, venues, themes, and authors.

We address five research questions:
\begin{enumerate}
\setlength{\itemsep}{0pt}
\setlength{\parskip}{0pt}
\setlength{\topsep}{0pt}
    \item How often are future work items in NLP addressed by later papers?
    \item How quickly do proposed directions receive follow-up?
    \item How does uptake vary by future work type and theme?
    \item How does follow-up move across publication venues?
    \item Is addressing driven mainly by the original authors or by the broader research community?
\end{enumerate}

Together, these questions treat future work as an empirical object for studying how stated research directions move through the field. Our results show that many future work items are later addressed, often within a few years, with clear differences across types, themes, venues, and author overlap.
%----------------------------------------------------------------------------------%

\section{Related Work}
\label{sec:related}
Scientific papers contain structured discourse signals that reveal how authors position their contributions, limitations, and open problems. Prior work has studied several such signals, including related work sections, peer reviews, limitations, and future work statements. Work on automatic related work summarization treats scholarly writing as a task of positioning a target paper within prior research \citep{hoang2010towards}, while work on review summarization and data paper linking shows how scientific documents can be connected through their argumentative, evaluative, and reuse functions \citep{uban2021generating,mcgillivray2025finding}. Closest to our setting, prior studies on future work mining extract and classify future work sentences as explicit descriptions of possible follow-up research \citep{hu2015mining,zhang2023automatic}. Recent work has also explored the related problem of generating future work sections and extracting or generating research limitations \citep{al2025futuregen,al2025bagels}. These studies establish future work and limitations as meaningful scholarly discourse units, but they mostly stop at extraction, classification, or generation. They do not ask whether the proposed directions are later pursued.

A separate line of work studies how scientific fields evolve over time. Topic modeling has been used to trace the history of ideas in computational linguistics, showing how topics rise, decline, and vary across venues \citep{hall2008studying}. Later work extends this view by examining topic relationships, including cooperation and competition between ideas \citep{tan2017friendships,uban2021studying}, and by showing that the rhetorical role assigned to a topic can predict its future growth or decline \citep{prabhakaran2016predicting}. These approaches provide important tools for studying research dynamics, but they usually operate at the level of topics, papers, or venues. As a result, they capture broad thematic movement rather than the fate of concrete research directions explicitly proposed by authors.

Citation-based studies offer a more direct view of scholarly uptake, but citations are also heterogeneous. Some citations indicate background, comparison, or acknowledgement, while others signal direct use or extension of prior work \citep{valenzuela2015identifying,cohan2019structural}. Citation intent and citation framing studies therefore show that scientific influence cannot be reduced to citation counts alone \citep{jurgens2018measuring}. Our work builds on this insight but shifts the unit of analysis from citations to future work statements. Instead of asking whether a paper was cited, we ask whether a specific proposed direction was addressed by later research. This connects scholarly discourse analysis with field-level research dynamics by treating future work statements as explicit proposed directions that can be tracked over time.
%----------------------------------------------------------------------------------%
\section{Dataset Construction}
\label{sec:construction}
We construct a dataset of future work statements from ACL Anthology papers and link each statement to later papers that may address it. The pipeline has four stages: source corpus construction, future work extraction, candidate follow-up retrieval, and addressing judgment. The resulting item-level labels support the temporal, thematic, venue-level, and author-level analyses in later sections.

\subsection{Source corpus}
We collect papers from the ACL Anthology published between 1979 and 2025, including ACL, EMNLP, NAACL, TACL, and related venues. For each paper, we parse metadata from the Anthology, including paper identifier, title, authors, year, venue, DOI, and available abstract. We download PDFs, convert them to plain text, and join the extracted future work items with paper metadata and reference lists. We retain papers with at least one extracted future work item and sufficient metadata for retrieval and linking, including publication year, title, venue, and either an abstract or usable converted text. Table \ref{tab:corpus_venue} reports the breakdown by venue.

\begin{table}[t]
\centering
\small
\begin{tabular}{lrrr}
\toprule
\textbf{Venue} & \textbf{Papers} & \textbf{FW items} & \textbf{FW / paper} \\
\midrule
ACL   &  9,999 & 30,570 & 3.1 \\
EMNLP &  8,808 & 28,218 & 3.2 \\
NAACL &  3,859 & 11,655 & 3.0 \\
TACL  &    447 &  1,627 & 3.6 \\
\midrule
Total & 23,113 & 72,070 & 3.1 \\
\bottomrule
\end{tabular}
\caption{Source corpus by venue.}
\label{tab:corpus_venue}
\end{table}

\subsection{Future work extraction}
\label{sec:extraction}
We follow prior work that treats future work as a sentence-level discourse signal that can be extracted and categorized from scientific articles \citep{hu2015mining,zhang2023automatic}. We identify future work items using three components: a heuristic extractor, a sentence-level SciBERT classifier \citep{beltagy2019scibert}, and an LLM-based annotation step. The heuristic and SciBERT components locate future-oriented text and support model development, and the LLM annotations provide the items used in the analyzed dataset.

The heuristic extractor uses section structure and lexical cues. It first identifies candidate sections whose headers match variants such as \textit{Future Work}, \textit{Conclusions and Future Work}, \textit{Limitations}, or \textit{Discussion}. Since many papers do not have a dedicated future work section, we also inspect conclusion sections and final paragraphs. Within these regions, the extractor flags sentences with future-oriented cues, including modal verbs such as \textit{will}, \textit{would}, \textit{could}, and \textit{should}; intention phrases such as \textit{we plan to}, \textit{we aim to}, and \textit{we intend to}; and fixed expressions such as \textit{in future work}, \textit{future research}, and \textit{an interesting direction}. It filters acknowledgements, funding statements, past-tense summaries, and generic paper summaries.

We then fine-tune SciBERT as a binary classifier for future-work versus non-future-work sentences. The SciBERT classifier is used as an intermediate filtering and calibration component, while the final released item spans and categories are produced by the LLM annotation step. Training examples come from three sources: positive examples from the heuristic extractor, hard negatives from introduction and abstract sentences of the same papers, and labeled examples from FutureGen dataset \citep{al2025futuregen}. The training set contains approximately 45,000 sentences, with about 60\% negative examples. We fine-tune with AdamW using a learning rate of $2{\times}10^{-5}$, batch size 32, two epochs, and maximum sequence length 128. At inference time, we use a probability threshold of 0.5, selected on a held-out set of 500 manually labeled sentences.

For the final corpus annotations, we use Gemini 2.5 Flash \citep{comanici2025gemini}. The model receives the conclusion and future-work sections of each paper, or the final 800 tokens when no relevant section boundary is detected. The full extraction prompt is provided in Appendix \ref{app:extraction-prompt}.

Each extracted item is assigned to one of four categories:
\begin{itemize}
\setlength{\itemsep}{0pt}
\setlength{\parskip}{0pt}
\setlength{\parsep}{0pt}
\setlength{\topsep}{0pt}
    \item \textbf{Limitation Fix}: addressing a weakness, failure case, or constraint of the current work;
    \item \textbf{Extension}: extending the method, data, task, language, domain, or evaluation setting;
    \item \textbf{New Direction}: proposing a new research path inspired by the current work;
    \item \textbf{Application}: using the method, data, or findings in a downstream system or real-world setting.
\end{itemize}

These categories describe the role of the proposed direction, not the topic of the paper. Table \ref{tab:fw_categories} reports the category distribution.

\begin{table}[t]
\centering
\small
\begin{tabular}{lrr}
\toprule
\textbf{Category} & \textbf{Count} & \textbf{\%} \\
\midrule
Limitation Fix  & 35,529 & 49.3 \\
Extension       & 21,800 & 30.2 \\
New Direction   & 10,643 & 14.8 \\
Application     &  4,096 &  5.7 \\
\midrule
Total           & 72,070 & 100.0 \\
\bottomrule
\end{tabular}
\caption{Distribution of future work items by category.}
\label{tab:fw_categories}
\end{table}

Together, these steps yield the released dataset of 72,070 future work items from 23,113 papers, each with a sentence span, a category label, and source paper metadata.

\section{Analysis Methods}
\label{sec:analysis-methods}
After extracting and categorizing future work items, we link them to later papers and derive the labels used in our uptake analyses. This section describes candidate retrieval, addressing judgments, and thematic clustering.

\subsection{Candidate follow-up retrieval}
\label{sec:retrieval}
To identify possible follow-up work, we combine citation-based and content-based retrieval. This follows prior evidence that citations are useful but incomplete signals of scholarly uptake, since not all citations indicate substantive use or extension, and relevant work may not cite the original paper \citep{valenzuela2015identifying}. For each future work item, we retrieve later papers that could plausibly address the proposed direction. In all cases, candidate papers must be published after the source paper.

The citation-based path uses direct citers of the source paper. We extract reference lists from the ACL Anthology and supplement them with OpenAlex metadata \citep{priem2022openalex} when available. For each source paper $p$, we scan papers $q$ published after $p$ and check whether $p$ appears in $q$'s reference list. Because reference metadata is noisy, we match references using word-token Jaccard similarity over normalized titles. We accept a reference match when similarity is at least 0.80, a threshold chosen empirically after comparing several values in preliminary experiments. Lower thresholds introduced many false matches between papers with overlapping title vocabulary, while higher thresholds missed legitimate references with minor title variation. Each matched citer becomes a candidate follow-up paper for all future work items from the source paper.

The global retrieval path captures follow-up work that may not cite the source paper. For each future work item, we build an augmented query by combining the future work statement with the source paper abstract. This grounds generic statements, such as \textit{we leave this to future work}, in the paper's topic. We encode the query and candidate abstracts with SPECTER \citep{cohan2020specter} and rank candidates by cosine similarity. We also compute BM25 similarity between the query and each candidate abstract. The final retrieval score combines semantic and lexical evidence:
\[
s = 0.7 \times s_{\text{semantic}} + 0.3 \times s_{\text{BM25}} .
\]
We retain candidates with $s \geq 0.45$ and pass at most the top $k=5$ candidates per future work item to the judgment step. These hyperparameters were chosen after preliminary tuning on a development sample. The similarity threshold filters out weakly related papers before LLM judgment, while the top-$k$ cap limits annotation cost and reduces repeated judgments over distant neighbours. In practice, increasing $k$ mainly added topically adjacent papers that rarely changed the addressed/unaddressed label.

\subsection{Addressing judgment}
\label{sec:judgment}
Candidate pairs are judged with Gemini 2.5 Flash \citep{comanici2025gemini}. Each input consists of a future work statement, source paper context, and candidate paper abstract. The prompt asks the model to perform two checks. First, it verifies whether the source and candidate papers are topically aligned. Second, if they are aligned, it decides whether the candidate paper addresses the specific future work direction.

The model returns one of three labels:
\begin{itemize}
\setlength{\itemsep}{0pt}
\setlength{\parskip}{0pt}
\setlength{\parsep}{0pt}
\setlength{\topsep}{0pt}
    \item \textbf{Yes}: the candidate carries out the task, extension, application, or limitation fix named in the future work statement;
    \item \textbf{Partial}: the candidate is adjacent to the direction or addresses part of it;
    \item \textbf{No}: the candidate is unrelated or topically nearby but does not engage the future work item.
\end{itemize}

Each judgment also includes a confidence label and a short explanation. The full prompt is provided in Appendix \ref{app:verification-prompt}. In the main analyses, we count \textit{yes} and \textit{partial} as addressed, since both indicate some uptake by later research. We use \textit{addressed} to mean that later work measurably engages the proposed direction, not that the underlying problem has been fully solved.

To estimate judgment quality, we draw a stratified validation sample of 200 future work items: 50 high-confidence addressed items, 50 lower-confidence addressed items, 50 unaddressed items, and 50 borderline cases where retrieval similarity is high but the model returns a negative verdict. This design covers both straightforward cases and difficult near-matches. The sample is manually annotated by two annotators who were not involved in writing the paper, allowing us to validate the LLM judgments and characterize common error types.

\subsection{Thematic clustering}
\label{sec:theme-clustering}
To analyze thematic variation in future work uptake, we cluster the extracted future work items. Each item is encoded with SPECTER \citep{cohan2020specter}, and we apply $k$-means clustering to the resulting embeddings. We select the number of clusters automatically using silhouette score over candidate values of $k$, which yields 14 clusters.

For each cluster, we extract representative keywords using TF-IDF over bigrams and trigrams. We then use Gemini 2.5 Flash \citep{comanici2025gemini} to assign a short human-readable label to each cluster, based on the cluster keywords and representative future work items. These labels are used only as descriptive summaries of induced clusters, and all quantitative analyses are computed over the cluster assignments themselves.

%----------------------------------------------------------------------------------%
\section{Experiments and Results}
\label{sec:results}
We analyze future work items at several levels: overall addressing rates, temporal dynamics, categories, venue-level patterns, thematic clusters, and author overlap. Unless otherwise stated, \textit{yes} and \textit{partial} LLM verdicts are counted as addressed. Because \textit{partial} labels indicate measurable continuation but weaker evidence than \textit{yes} labels, we combine both labels and treat the resulting rates as approximate. Items for which the pipeline retrieved no candidate (16.3\% of all FW items) are treated as not addressed by default. Where multiple candidates were retrieved for a single FW item, we apply a union rule: the item counts as addressed if any candidate received a \textit{yes} or \textit{partial} verdict.

\subsection{How often is future work addressed?}
\label{sec:results-overall}
\paragraph{Human validation.}
We validate the addressing judgments on a diagnostic sample of 200 future work items. The sample is split into four groups based on the LLM output and retrieval score: high-confidence addressed items, lower-confidence addressed items, unaddressed items, and high-similarity borderline negatives. Each group contains 50 items. Two annotators who were not involved in writing the paper independently label each item as \textit{addressed} or \textit{not addressed}. Inter-annotator agreement is fair ($\kappa=0.38$), indicating that the addressing decision is difficult, especially for near-matches.

We report LLM agreement against both annotators. Against the more conservative annotator, LLM-human agreement is substantial ($\kappa=0.72$), with 86.0\% binary accuracy, 73.0\% precision, and 83.9\% F1 for the addressed class. Against the second annotator, agreement is moderate ($\kappa=0.55$). These scores are diagnostic validation results, not corpus-level performance estimates, because the sample is balanced across LLM-defined groups and not sampled in proportion to their corpus prevalence.

The largest disagreement appears in the high-similarity borderline group: the conservative annotator labels 0\% of these items as addressed, while the second annotator labels 46\% as addressed. This suggests that borderline cases are especially sensitive to annotation criteria. The LLM also over-credits adjacent work as \textit{partial} in the positive groups. We therefore treat the 50.0\% corpus-level addressing rate as an approximate pipeline estimate, not as a precise prevalence estimate. As a conservative sensitivity check, multiplying the detected corpus-level rate by the observed precision against the conservative annotator gives an estimate of approximately 37\%. Table \ref{tab:validation} summarizes the validation results.

\begin{table}[t]
\centering
\small
\begin{tabular}{lccccc}
\toprule
\textbf{Group} & \textbf{n} & \textbf{LLM} & \textbf{A} & \textbf{B} & \textbf{Acc. A} \\
\midrule
Addr. (high) & 50 & 100\% & 78\% & 90\% & 78\% \\
Addr. (low)  & 50 & 100\% & 68\% & 68\% & 68\% \\
Borderline   & 50 &   0\% &  0\% & 46\% & 100\% \\
Unaddr.      & 50 &   0\% &  2\% &  2\% & 98\% \\
\midrule
\textbf{Overall} & 200 & 50\% & 37\% & 52\% & 86\% \\
\bottomrule
\end{tabular}
\caption{Diagnostic validation results by LLM-defined group. LLM, A, and B indicate the percentage of items labeled as addressed. A is the more conservative annotator. Acc. A is binary accuracy against annotator A.}
\label{tab:validation}
\end{table}

Across the full corpus, 50.0\% of future work items are detected as addressed by at least one later paper. Given the validation results, we treat this as substantial but noisy evidence, not as a precise prevalence estimate. Addressing rates vary across venues. TACL has the highest rate, with 57.0\% of future work items addressed, followed by ACL at 51.7\%, EMNLP at 49.9\%, and NAACL at 44.7\%. These differences should not be interpreted as venue quality rankings. They may reflect topic composition, publication volume, paper length, citation practices, and the time available for follow-up work to appear. Table \ref{tab:venue_addressing} reports the venue-level breakdown.

\begin{table}[t]
\centering
\small
\begin{tabular}{lrrr}
\toprule
\textbf{Venue} & \textbf{FW items} & \textbf{Addressed} & \textbf{Rate} \\
\midrule
ACL   & 30,570 & 15,793 & 51.7\% \\
EMNLP & 28,218 & 14,092 & 49.9\% \\
NAACL & 11,655 &  5,205 & 44.7\% \\
TACL  &  1,627 &    927 & 57.0\% \\
\midrule
Total & 72,070 & 36,017 & 50.0\% \\
\bottomrule
\end{tabular}
\caption{Addressing rates by venue.}
\label{tab:venue_addressing}
\end{table}

\subsection{How quickly is future work addressed?}
\label{sec:results-time}
Among items detected as addressed, the median time to address is two years and the mean is 4.85 years. Follow-up often appears quickly: 62.9\% of addressed items are followed up within two years, and 78.8\% within five years. At the same time, the distribution has a long tail. Some directions are addressed more than ten years after they are proposed, and a small number of extreme lags may reflect errors in year attribution rather than genuine latency. Time is measured as the integer difference between the addressing paper's publication year and the source paper's publication year. Same-year pairs are excluded from the distribution because publication order within a year is often ambiguous in Anthology metadata.

This pattern suggests two modes of follow-up. Many future work statements describe near-term opportunities that can be pursued quickly, often because the task, dataset, or method is already well defined. Other statements may be addressed only after new data, computational resources, evaluation practices, or research priorities make them tractable.

Temporal comparisons require caution because recent papers have had less time to accumulate follow-up work. This right-censoring is especially important for 2024 and 2025. We therefore treat recent-year addressing rates as lower-bound estimates, not as evidence that current future work is less likely to be pursued. Some earlier cohorts also show anomalously low rates in the pipeline output, so we avoid drawing substantive conclusions from isolated year-level fluctuations.

\begin{table}[t]
\centering
\small
\begin{tabular}{cc}
\toprule
\textbf{Years after publication} & \textbf{Cumulative \%} \\
\midrule
$\leq$ 1 year  & 47.7 \\
$\leq$ 2 years & 62.9 \\
$\leq$ 3 years & 71.1 \\
$\leq$ 4 years & 75.7 \\
$\leq$ 5 years & 78.8 \\
$\leq$ 10 years & 81.7 \\
$>$ 10 years & 18.3 \\
\bottomrule
\end{tabular}
\caption{Cumulative time to address among addressed future work items.}
\label{tab:time_to_address}
\end{table}

\subsection{Which types of future work are addressed?}
\label{sec:results-category}
Addressing rates differ substantially by future work category. Application-oriented directions are addressed most often, with a rate of 65.8\%. Extensions and new directions have similar rates, at 53.8\% and 53.7\%, respectively. Limitation Fix statements are the most common category, but have the lowest addressing rate, 44.7\%.

This contrast is informative because limitation statements and application statements play different roles in scholarly writing. Application items often describe concrete use cases or downstream settings, making them easier for later work to operationalize. Limitation Fix items, by contrast, may point to harder technical barriers, known weaknesses of a paradigm, or problems that are acknowledged but not immediately attractive as standalone contributions.

\begin{table}[t]
\centering
\small
\begin{tabular}{lrrr}
\toprule
\textbf{Category} & \textbf{FW items} & \textbf{Addressed} & \textbf{Rate} \\
\midrule
Limitation Fix & 35,529 & 15,884 & 44.7\% \\
Extension      & 21,800 & 11,727 & 53.8\% \\
New Direction  & 10,643 &  5,712 & 53.7\% \\
Application    &  4,096 &  2,694 & 65.8\% \\
\midrule
Total          & 72,070 & 36,017 & 50.0\% \\
\bottomrule
\end{tabular}
\caption{Addressing rates by future work category.}
\label{tab:category_addressing}
\end{table}

The 21-point gap between Application and Limitation Fix items suggests that the form in which authors phrase future work matters. Directions framed as concrete applications appear more likely to be taken up than directions framed as weaknesses to be repaired.

\subsection{How does follow-up move across venues?}
\label{sec:results-crossvenue}
We next examine venue cross-pollination: for each addressed item, we compare the venue of the source paper with the venue of the addressing paper. When multiple candidate papers address the same item, we attribute it to the venue of the earliest addressing paper (ties broken by the highest LLM confidence score). Table \ref{tab:venue_crosspollination} shows that future work uptake is not confined within venue boundaries. ACL and EMNLP form the strongest bidirectional pair: 5,367 ACL future work items are addressed by EMNLP papers, while 4,578 EMNLP items are addressed by ACL papers.

TACL shows a different pattern. Although TACL has the highest addressing rate overall, most follow-up to TACL future work appears outside TACL, especially in ACL and EMNLP. This is expected given TACL's smaller publication volume and broader journal-style role in the field. NAACL future work is also more often addressed by ACL and EMNLP than by later NAACL papers.

\begin{table*}[t]
\centering
\small
\begin{tabular}{lrrrr}
\toprule
\textbf{Source / Addressing} & \textbf{ACL} & \textbf{EMNLP} & \textbf{NAACL} & \textbf{TACL} \\
\midrule
ACL   & \textbf{5,645} & 5,367 & 2,723 & 2,058 \\
EMNLP & 4,578 & \textbf{4,948} & 2,285 & 2,281 \\
NAACL & 1,903 & 1,743 & \textbf{734} & 825 \\
TACL  & 319 & 327 & 147 & \textbf{134} \\
\bottomrule
\end{tabular}
\caption{Venue cross-pollination. Rows: venue of the source paper; columns: venue of the addressing paper. Bold diagonal values show within-venue follow-up.}
\label{tab:venue_crosspollination}
\end{table*}

These results support the view that future work statements can function as field-level signals. They are not only followed up within the same venue community, they circulate across the major publication venues of NLP.

\begin{table*}[t]
\centering
\small
\begin{tabular}{lrr}
\toprule
\textbf{Cluster theme} & \textbf{Items} & \textbf{Addr. rate} \\
\midrule
Predicate-Argument Structure Disambiguation & 4,653 & 62.5\% \\
Cross-lingual transfer learning             & 5,629 & 59.5\% \\
Domain-specific summarization               & 5,313 & 59.1\% \\
Cross-lingual syntactic transfer            & 5,544 & 58.4\% \\
Multi-turn dialogue generalization          & 3,928 & 55.2\% \\
Knowledge Graph Enrichment                  & 4,842 & 52.1\% \\
Cross-domain reasoning generalization       & 3,315 & 51.2\% \\
Efficient LLM Fine-tuning                   & 6,342 & 47.9\% \\
Multimodal Vision-Language Alignment        & 3,432 & 45.0\% \\
Cross-cultural bias detection               & 4,154 & 44.1\% \\
Human-in-the-loop evaluation                & 4,301 & 43.3\% \\
Scalable Clustering Algorithms              & 6,951 & 41.6\% \\
Efficient Software Engineering Optimization & 4,716 & 37.8\% \\
Adversarial attack defense                  & 2,413 & 37.3\% \\
\bottomrule
\end{tabular}
\caption{Addressing rates by thematic cluster, sorted by addressing rate.}
\label{tab:theme_addressing}
\end{table*}

\subsection{Which themes receive the most follow-up?}
\label{sec:results-themes}
Using the clustering procedure described in Section \ref{sec:theme-clustering}, we group future work items into 14 thematic clusters. The clustering covers 65,533 of the 72,070 extracted items (91.0\%); the remaining 6,537 items are excluded because they fall below the minimum length threshold used before embedding. The resulting labels and addressing rates are reported in Table \ref{tab:theme_addressing}. Addressing rates vary strongly across themes. Predicate-Argument Structure Disambiguation has the highest addressing rate at 62.5\%, followed by Cross-lingual transfer learning at 59.5\%, Domain-specific summarization at 59.1\%, and Cross-lingual syntactic transfer at 58.4\%. 

These high-addressing themes share several properties: they are established areas of NLP, often have clear evaluation settings, and have attracted sustained community attention. In contrast, Adversarial attack defense and Efficient Software Engineering Optimization have the lowest rates, at 37.3\% and 37.8\%, respectively. These directions may be harder to address because they involve moving targets, infrastructure constraints, or rapidly changing technical standards. Because the labels are LLM-generated summaries of induced clusters, we interpret them as descriptive aids rather than fixed ontological categories. Individual clusters may contain heterogeneous future work items that share embedding-level similarity but differ in specific task, method, or domain. Appendix \ref{app:cluster-examples} gives representative examples from each cluster.

The cluster results show that follow-up is not only a property of individual statements. It also appears related to field-level infrastructure. Active communities, shared datasets, stable benchmarks, and available computational resources may make some proposed directions easier to pursue than others.

\subsection{Who addresses future work?}
\label{sec:selfvscommunity}
We classify each addressed FW item as \textit{self-addressed} if at least one original author of the source paper is also an author on the addressing paper, and \textit{community-addressed} if there is no author overlap. Author names are compared after Unicode normalization and accent stripping (NFD decomposition), so variant spellings such as \textit{Vuli\'{c}} and \textit{Vulic} are treated as the same author.

Most addressed items are followed up by papers with no detected author overlap: 93.1\% are classified as community-addressed, 5.3\% as self-addressed, and 1.6\% as ambiguous or partial author matching cases. Since author matching relies on normalized surface names, the self-addressing rate should be interpreted as an approximate lower-bound estimate.

This result suggests that future work statements often function less as private author plans and more as public signals of open research opportunities. Once stated in a paper, a future direction becomes available to the broader community, which may pursue it independently of the original authors.

The low self-addressing rate is especially notable for Limitation Fix items. One might expect authors to be well positioned to address limitations of their own work, but the aggregate pattern indicates that even these items are usually pursued by other researchers. This does not imply that authors do not intend to follow their own proposed directions. It reflects the distributed nature of NLP research, where many groups work on related problems in parallel and where follow-up work can emerge from outside the original author team.

Author-level uptake is uneven, but we treat these patterns descriptively because they are affected by productivity, co-authorship structure, topic popularity, and time in the field. We report illustrative author-level rankings in Appendix \ref{app:author-rankings}.

%----------------------------------------------------------------------------------%
\section{Discussion}
\label{sec:discussion}
Our results suggest that future work statements are more than a conventional way to close a paper. Many extracted items are later connected to subsequent research, although the validation results show that the exact rate depends on annotation criteria. We therefore interpret these statements as useful but noisy evidence for how research directions develop.

The results also show that not all proposed directions are equally easy to pursue. Application-oriented items are detected as addressed more often than limitation fixes, even though limitation fixes are the most common type. This pattern suggests that framing matters. Applications often name a concrete task, dataset, domain, or use case. Limitation fixes may point to harder technical barriers or to weaknesses that are acknowledged but not easy to turn into standalone contributions.

The strong community-addressing pattern changes how these sections can be interpreted. Most addressed items are followed up by papers with no detected author overlap. This does not mean that authors do not intend to pursue their own proposed directions. It suggests that, once published, these statements can enter a shared research space where other groups may respond to, extend, or repurpose them.

The thematic and venue-level analyses point to the role of field-level infrastructure. Directions in established areas with shared benchmarks, active communities, and clear evaluation protocols are detected as addressed more often. Themes involving moving targets, expensive infrastructure, or less stable evaluation practices receive less measurable follow-up in our pipeline. This should not be read as evidence that low-rate themes are less important. Some may expose difficult methodological, social, or computational challenges.

Finally, our analysis shows the limits of treating future work statements as data. The items are often brief, heterogeneous, and sometimes vague. A later paper may partially address a direction, address it without citing the source, or use different terminology for a similar idea. For this reason, the dataset is best used to study broad patterns, not to decide whether any individual statement was definitively solved.
%----------------------------------------------------------------------------------%
\section{Conclusions}
We presented a dataset and analysis pipeline for studying what happens to future work statements in NLP research. Instead of treating future work only as a textual convention, we treat it as a set of explicit proposed directions that can be tracked in later work. Using 72,070 future work items from 23,113 ACL Anthology papers, we link proposed directions to later candidate papers and estimate whether they are addressed.

Our analyses show that many future work items are later addressed, but not evenly. Many receive follow-up within a few years, while others remain unaddressed or are taken up only after a longer delay. Application-oriented directions are detected as addressed more often than limitation fixes, and thematic variation suggests that shared benchmarks, active communities, and available infrastructure shape which proposed directions are easier to pursue. We also find that follow-up is largely community-driven: most addressed items are taken up by researchers outside the original author team.

These findings suggest that future work statements can support a broader empirical view of how research directions develop. They are not definitive records of whether an idea has been solved, but they offer a useful signal of how researchers formulate open opportunities and how those opportunities later circulate through the field. 
%----------------------------------------------------------------------------------%
\section*{Limitations}
Our study has several limitations. The dataset depends on automatic extraction of future work items. Although the extraction prompt is designed to select concrete and self-contained proposed directions, some extracted items may be vague, incomplete, or not genuine future work statements. This affects the denominator of all addressing estimates. A larger manual audit of extraction quality would further strengthen the resource.

Addressing judgments are also difficult. A later paper may address only part of a proposed direction, use different terminology, or pursue the same idea without citing the source paper. Human validation shows that the addressed relation is sensitive to annotation criteria, especially for near-matches. For this reason, the reported 50.0\% detected addressing rate should be interpreted as an approximate pipeline estimate. A conservative sensitivity check gives an estimate of approximately 37\%.

The retrieval step limits what the judgment step can observe. If no relevant candidate paper is retrieved, an item is treated as not addressed, which may underestimate follow-up for directions addressed by papers outside the retrieved set. Broad or vague future work items may also retrieve topically related papers that are not true follow-up work.

Temporal, author-level, and thematic analyses should be read descriptively. Recent papers, especially those published in 2024 and 2025, have had little time to accumulate follow-up work. Older papers may be affected by incomplete metadata, noisy PDF conversion, and errors in publication year attribution. The self-versus-community analysis relies on normalized surface names instead of full author disambiguation, so the self-addressing rate should be treated as an approximate lower-bound estimate. Finally, thematic clusters are induced from embeddings and labeled with an LLM. They are useful for broad patterns, but they should not be read as fixed categories of NLP research.

%----------------------------------------------------------------------------------%
\section*{Ethical Considerations}
This work analyzes published scientific papers and scholarly metadata. It does not use private data or human-subject data. The pipeline is automatic, so it can make mistakes, especially for vague statements, noisy PDF text, incomplete metadata, and borderline cases. We therefore report broad patterns, not final judgments about individual papers, authors, venues, or research areas. Author-level results are descriptive only and should not be used for evaluation, hiring, funding, or other consequential decisions. We release the dataset and will release the pipeline with documentation on label quality, known errors, and appropriate uses.

%----------------------------------------------------------------------------------%

\section*{Acknowledgments}
This work was partially supported by the project “Romanian Hub for Artificial Intelligence - HRIA”, Smart Growth, Digitization and Financial Instruments Program, 2021-2027, MySMIS no. 351416 and a grant of the Romanian Ministry of Research, Innovation and Digitalization, CNCS - UEFISCDI (Center of Excellence for Climate and Societal Change), project number PN-IV-P6-6.1-CoEx-2024-004, within PNCDI IV and a grant of the Ministry of Education and Research, CNCS-UEFISCDI, project SIROLA, number PN-IV-P1- PCE-20231701, within PNCDI IV.

%----------------------------------------------------------------------------------%
% \section*{Acknowledgments}
% \blindtext
\bibliography{custom}

@inproceedings{hoang2010towards,
  title={Towards automated related work summarization},
  author={Hoang, Cong Duy Vu and Kan, Min-Yen},
  booktitle={Coling 2010: Posters},
  pages={427--435},
  year={2010}
}

@article{uban2021generating,
  title={Generating Summaries for Scientific Paper Review},
  author={Uban, Ana Sabina and Caragea, Cornelia},
  journal={arXiv preprint arXiv:2109.14059},
  year={2021}
}

@inproceedings{mcgillivray2025finding,
  title={Finding the Paper Behind the Data: Automatic Identification of Research Articles related to Data Publications},
  author={McGillivray, Barbara and Aryan, Kaveh and Harperath, Viola and Ribary, Marton and Wigdorowitz, Mandy},
  booktitle={Proceedings of the Third Workshop for Artificial Intelligence for Scientific Publications},
  pages={34--43},
  year={2025}
}

@article{hu2015mining,
  title={Mining and analyzing the future works in scientific articles},
  author={Hu, Yue and Wan, Xiaojun},
  journal={arXiv preprint arXiv:1507.02140},
  year={2015}
}

@article{zhang2023automatic,
  title={Automatic recognition and classification of future work sentences from academic articles in a specific domain},
  author={Zhang, Chengzhi and Xiang, Yi and Hao, Wenke and Li, Zhicheng and Qian, Yuchen and Wang, Yuzhuo},
  journal={Journal of Informetrics},
  volume={17},
  number={1},
  pages={101373},
  year={2023},
  publisher={Elsevier}
}

@inproceedings{al2025futuregen,
  title={FutureGen: A RAG-based Approach to Generate the Future Work of Scientific Article},
  author={Al Azher, Ibrahim and Mokarrama, Miftahul Jannat and Guo, Zhishuai and Choudhury, Sagnik Ray and Alhoori, Hamed},
  booktitle={2025 IEEE International Conference on eScience (eScience)},
  pages={427--438},
  year={2025},
  organization={IEEE}
}

@article{al2025bagels,
  title={BAGELS: Benchmarking the Automated Generation and Extraction of Limitations from Scholarly Text},
  author={Al Azher, Ibrahim and Mokarrama, Miftahul Jannat and Guo, Zhishuai and Choudhury, Sagnik Ray and Alhoori, Hamed},
  journal={arXiv preprint arXiv:2505.18207},
  year={2025}
}

@inproceedings{hall2008studying,
  title={Studying the history of ideas using topic models},
  author={Hall, David and Jurafsky, Dan and Manning, Christopher D},
  booktitle={Proceedings of the 2008 conference on empirical methods in natural language processing},
  pages={363--371},
  year={2008}
}

@inproceedings{tan2017friendships,
  title={Friendships, rivalries, and trysts: Characterizing relations between ideas in texts},
  author={Tan, Chenhao and Card, Dallas and Smith, Noah A},
  booktitle={Proceedings of the 55th Annual Meeting of the Association for Computational Linguistics (Volume 1: Long Papers)},
  pages={773--783},
  year={2017}
}

@inproceedings{uban2021studying,
  title={Studying the evolution of scientific topics and their relationships},
  author={Uban, Ana Sabina and Caragea, Cornelia and Dinu, Liviu P},
  booktitle={Findings of the association for computational linguistics: acl-ijcnlp 2021},
  pages={1908--1922},
  year={2021}
}

@inproceedings{prabhakaran2016predicting,
  title={Predicting the rise and fall of scientific topics from trends in their rhetorical framing},
  author={Prabhakaran, Vinodkumar and Hamilton, William L and McFarland, Dan and Jurafsky, Dan},
  booktitle={Proceedings of the 54th Annual Meeting of the Association for Computational Linguistics (Volume 1: Long Papers)},
  pages={1170--1180},
  year={2016}
}

@inproceedings{valenzuela2015identifying,
  title={Identifying Meaningful Citations.},
  author={Valenzuela, Marco and Ha, Vu and Etzioni, Oren},
  booktitle={AAAI workshop: Scholarly big data},
  volume={15},
  pages={13},
  year={2015}
}

@inproceedings{cohan2019structural,
  title={Structural scaffolds for citation intent classification in scientific publications},
  author={Cohan, Arman and Ammar, Waleed and Van Zuylen, Madeleine and Cady, Field},
  booktitle={Proceedings of the 2019 conference of the North American chapter of the Association for Computational Linguistics: human language technologies, volume 1 (long and short papers)},
  pages={3586--3596},
  year={2019}
}

@article{jurgens2018measuring,
  title={Measuring the evolution of a scientific field through citation frames},
  author={Jurgens, David and Kumar, Srijan and Hoover, Raine and McFarland, Dan and Jurafsky, Dan},
  journal={Transactions of the Association for Computational Linguistics},
  volume={6},
  pages={391--406},
  year={2018}
}

@inproceedings{beltagy2019scibert,
  title={SciBERT: A pretrained language model for scientific text},
  author={Beltagy, Iz and Lo, Kyle and Cohan, Arman},
  booktitle={Proceedings of the 2019 conference on empirical methods in natural language processing and the 9th international joint conference on natural language processing (EMNLP-IJCNLP)},
  pages={3615--3620},
  year={2019}
}

@article{comanici2025gemini,
  title={Gemini 2.5: Pushing the frontier with advanced reasoning, multimodality, long context, and next generation agentic capabilities},
  author={Comanici, Gheorghe and Bieber, Eric and Schaekermann, Mike and Pasupat, Ice and Sachdeva, Noveen and Dhillon, Inderjit and Blistein, Marcel and Ram, Ori and Zhang, Dan and Rosen, Evan and others},
  journal={arXiv preprint arXiv:2507.06261},
  year={2025}
}

@inproceedings{cohan2020specter,
  title={Specter: Document-level representation learning using citation-informed transformers},
  author={Cohan, Arman and Feldman, Sergey and Beltagy, Iz and Downey, Doug and Weld, Daniel S},
  booktitle={Proceedings of the 58th annual meeting of the association for computational linguistics},
  pages={2270--2282},
  year={2020}
}

@article{priem2022openalex,
  title={OpenAlex: A fully-open index of scholarly works, authors, venues, institutions, and concepts},
  author={Priem, Jason and Piwowar, Heather and Orr, Richard},
  journal={arXiv preprint arXiv:2205.01833},
  year={2022}
}
%----------------------------------------------------------------------------------%
\appendix
\section{LLM Extraction Prompt}
\label{app:extraction-prompt}
We use the following prompt for extracting future work items from paper text. The model receives either the detected conclusion and future-work sections or, when no relevant section boundary is found, the final 800 tokens of the converted paper text. The prompt asks the model to return atomic future work items using a structured JSON schema. Each item contains a \texttt{sentence} field, preferably a verbatim span from the paper, and a \texttt{category} field selected from the four-way taxonomy described in Section \ref{sec:extraction}.

\begin{lstlisting}[style=promptstyle]
        You are a strict Data Curation Expert. 
        I will provide a list of sentences extracted from the Discussion/Conclusion section of an academic paper.
        
        Your Goal: Extract ONLY sentences that explicitly describe specific planned future work or unaddressed limitations the authors intend to solve.
        
        Input Candidates:
        \{candidate\_text\}
        
        --- GUIDELINES ---
        
        1. THE "VERB TENSE" TEST (CRITICAL):
           - DISCARD sentences using Past Tense verbs describing the current study (e.g., "We *proposed*...", "We *analyzed*...", "The model *achieved*...").
           - DISCARD Present Tense descriptions of results (e.g., "Figure 3 *shows*...", "Our method *outperforms*...").
           - KEEP sentences with Future Intent (e.g., "We *will* explore...", "We *plan* to...", "We *intend* to...").
           - KEEP sentences with Potential/Modal verbs (e.g., "It *remains* to be seen...", "Future work *could* address...", "A promising direction *is*...").

        2. NOISE FILTERING:
           - Aggressively discard Acknowledgments ("Supported by Grant X...").
           - Discard Figure/Table captions or references ("See Table 2...").
           - Discard general fluff ("Future work is needed" without saying WHAT work).

        3. CATEGORIZATION:
           - "Extension": Applying the method to new data/languages/domains.
           - "Limitation Fix": Solving a specific problem found in this study.
           - "Application": Using the model for a downstream task.
           - "New Direction": A completely new idea inspired by this work.

        Output strictly using the JSON schema provided.
\end{lstlisting}

%----------------------------------------------------------------------------------%
\section{LLM Verification Prompt}
\label{app:verification-prompt}
We use the following prompt to verify whether retrieved candidate papers address specific future work statements. The prompt is intentionally conservative: it first checks topical alignment between the source and candidate papers, then checks whether the candidate performs the concrete action requested by the future work statement. The model returns a verdict, confidence label, and short explanation for each pair.

\begin{lstlisting}[style=promptstyle]
You are an expert in academic research analysis. Your job is to decide, for each numbered pair, whether the candidate paper actually addresses the specific future-work statement made by the source paper.
 
You are a STRICT grader. Most candidate papers in this set do NOT genuinely address the source's future work, even when they share a research area. Topical relatedness is common; actually doing the called-for work is rare. When in doubt, choose the lower label. Reserve "yes" for cases that clearly clear the bar in both checks below.
 
Make TWO checks, in order, before giving a verdict.
 
CHECK 1 — Topic alignment
Are the source paper and candidate paper working on the same kind of problem in the same domain?
- Shared surface vocabulary is NOT alignment. Words like "robustness", "evaluation", "multilingual", "attention", "benchmark", "LLM", "bias", or "abstract" appearing in both papers count for nothing on their own.
- A shared method name applied to different problems is NOT alignment (e.g. both use "constrained decoding" but one is for parsing and one is for chart generation → fail).
- A shared object studied for different purposes is NOT alignment (e.g. both mention "protein-protein interactions" but one extracts them from text and one predicts them from sequences → fail).
Ask: would a researcher in the source paper's subfield consider the candidate part of their own line of work? If not, CHECK 1 fails.
If CHECK 1 fails, the verdict is "no". Do not proceed to CHECK 2.
 
CHECK 2 — Direct addressing of the future work
Does the candidate paper actually DO the specific thing the source paper called for — not merely operate near it?
Distinguish three cases:
- ADDRESSES: the candidate performs the concrete task, extension, or fix named in the future-work statement. This earns "yes".
- ADJACENT: the candidate works on the same topic and is in the neighborhood of the future-work direction, but does not do the specific thing named, OR does a related-but-different thing, OR only enables/motivates the work rather than carrying it out. This earns "partial".
- UNRELATED TO THE FW: the candidate does not engage the future-work direction at all (even if CHECK 1 passed). This earns "no".
Test yourself: name the exact action the future-work statement requests, then point to the sentence in the candidate abstract that performs that exact action. If you cannot point to it, it is at most "partial".
 
DEGENERATE INPUTS — read carefully
Some inputs are missing or uninformative. Handle them conservatively rather than guessing:
- If the future-work statement is vague, generic, or contentless (e.g. "we leave this to future work", "these are interesting directions", "performance could be improved with a different model", "this work can be extended in several ways"), there is no specific work to match. The verdict cannot be "yes". It is "partial" only if CHECK 1 clearly passes AND the candidate plausibly continues the source's line; otherwise "no". Set confidence to "low" or "medium".
- If the source abstract is only a title or is too truncated to determine the research problem, do not infer a rich topic from the title alone. Judge only on what is verifiable; default to "no" unless alignment and addressing are both still clearly evident. Lower your confidence accordingly.
- If text appears garbled (OCR noise) and you cannot recover the meaning, do not award credit for words you cannot interpret.
 
VERDICT RULES (apply strictly)
- "yes"     — CHECK 1 passes AND CHECK 2 is ADDRESSES (the specific future work is actually carried out).
- "partial" — CHECK 1 passes AND CHECK 2 is ADJACENT (same topic, related to the direction, but the specific work is not actually done).
- "no"      — CHECK 1 fails, OR CHECK 2 is UNRELATED, OR the match rests on shared vocabulary/method-name/object only, OR the future-work statement is too vague to match against and the link is not clearly continuous.
 
CONFIDENCE
- "high"   — abstracts are informative and the call is clear-cut.
- "medium" — some inference required, or the partial/yes boundary is genuinely close.
- "low"    — degenerate input (title-only source, vague FW, garbled text) forced a judgment with weak evidence.
 
Reply with a JSON array of objects, one per pair, in the same order:
[{{"id": 1, "addressed": "yes"|"partial"|"no", "confidence": "high"|"medium"|"low", "explanation": "..."}}, ...]
 
In the explanation: (1) state whether CHECK 1 passed or failed and why (name the shared problem, or name the vocabulary/method/object that only superficially overlaps); (2) if CHECK 1 passed, name the exact action the future-work statement requests and state whether the candidate ADDRESSES or is only ADJACENT to it. Keep it to 1-3 sentences.
\end{lstlisting}

\section{Thematic Cluster Examples}
\label{app:cluster-examples}

Table~\ref{tab:theme_addressing} reports induced clusters whose labels are descriptive summaries. Below, we show two representative items from each cluster. The examples are lightly cleaned for readability and illustrate the range of items within each cluster. 

\paragraph{Predicate argument structure disambiguation (62.5\%)}
\begin{itemize}
\setlength{\itemsep}{0pt}
\setlength{\parskip}{0pt}
\setlength{\topsep}{0pt}
\item \textit{We plan to expand our experiments to more languages as well as NomBank-style noun labels.}
\item \textit{Lexical algorithms that ignore predicate-argument structure perform worse in the presence of quantifiers, which we identify as a direction for future work.}
\end{itemize}

\paragraph{Cross-lingual transfer learning (59.5\%)}
\begin{itemize}
\setlength{\itemsep}{0pt}
\setlength{\parskip}{0pt}
\setlength{\topsep}{0pt}
\item \textit{Future work will extend this approach to other language families and different multilingual tasks, such as machine translation.}
\item \textit{Future work should address the lack of diverse data for target languages beyond English.}
\end{itemize}

\paragraph{Domain-specific summarization (59.1\%)}
\begin{itemize}
\setlength{\itemsep}{0pt}
\setlength{\parskip}{0pt}
\setlength{\topsep}{0pt}
\item \textit{While domain-specific long-text datasets are currently limited, we recognize their importance and will explore this direction in future work.}
\item \textit{Scaling the dataset beyond its current size is a clear next step for future work.}
\end{itemize}

\paragraph{Cross-lingual syntactic transfer (58.4\%)}
\begin{itemize}
\setlength{\itemsep}{0pt}
\setlength{\parskip}{0pt}
\setlength{\topsep}{0pt}
\item \textit{Future work could explore other linguistic features, such as context-free grammar phrase structure rules or tree substitution grammars.}
\item \textit{We plan to extend these models to other tasks such as syntactic parsing and machine translation.}
\end{itemize}

\paragraph{Multi-turn dialogue generalization (55.2\%)}
\begin{itemize}
\setlength{\itemsep}{0pt}
\setlength{\parskip}{0pt}
\setlength{\topsep}{0pt}
\item \textit{Further research is needed to expand to more complex tasks such as conversational or dialogue-based prompting.}
\item \textit{Future work should test whether these findings generalize to other base and chat models.}
\end{itemize}

\paragraph{Knowledge graph enrichment (52.1\%)}
\begin{itemize}
\setlength{\itemsep}{0pt}
\setlength{\parskip}{0pt}
\setlength{\topsep}{0pt}
\item \textit{The knowledge base can be enriched by adding more relations from a variety of linguistic resources, including co-occurrence relations from large corpora.}
\item \textit{Future work will continue expanding the resource by increasing data diversity and incorporating multilingual annotations.}
\end{itemize}

\paragraph{Cross-domain reasoning generalization (51.2\%)}
\begin{itemize}
\setlength{\itemsep}{0pt}
\setlength{\parskip}{0pt}
\setlength{\topsep}{0pt}
\item \textit{Future work should include a more comprehensive evaluation across reasoning and non-reasoning tasks.}
\item \textit{The framework could be extended to include a broader range of hypotheses, fallacy types, data modalities, and reasoning tasks.}
\end{itemize}

\paragraph{Efficient LLM fine-tuning (47.9\%)}
\begin{itemize}
\setlength{\itemsep}{0pt}
\setlength{\parskip}{0pt}
\setlength{\topsep}{0pt}
\item \textit{Future work may require using larger models or designing new training schemes.}
\item \textit{Improving the quality of generated training samples remains an important direction for future work.}
\end{itemize}

\paragraph{Multimodal vision-language alignment (45.0\%)}
\begin{itemize}
\setlength{\itemsep}{0pt}
\setlength{\parskip}{0pt}
\setlength{\topsep}{0pt}
\item \textit{The techniques developed in this work can be extended to other domains and modalities.}
\item \textit{Future work should examine the reliability of cross-modal alignments under distribution shift.}
\end{itemize}

\paragraph{Cross-cultural bias detection (44.1\%)}
\begin{itemize}
\setlength{\itemsep}{0pt}
\setlength{\parskip}{0pt}
\setlength{\topsep}{0pt}
\item \textit{The method can be used to compare biases within and across cultural groups and genres.}
\item \textit{Future work should study many more languages and cultures before drawing global conclusions.}
\end{itemize}

\paragraph{Human-in-the-loop evaluation (43.3\%)}
\begin{itemize}
\setlength{\itemsep}{0pt}
\setlength{\parskip}{0pt}
\setlength{\topsep}{0pt}
\item \textit{We leave the development of a better evaluation framework specifically tailored to instructions for future work.}
\item \textit{Future work should investigate how human feedback can be incorporated into evaluation protocols.}
\end{itemize}

\paragraph{Scalable clustering algorithms (41.6\%)}
\begin{itemize}
\setlength{\itemsep}{0pt}
\setlength{\parskip}{0pt}
\setlength{\topsep}{0pt}
\item \textit{Extending our investigation to larger settings would provide more reliable results.}
\item \textit{Future work should address the difficulties of rigorous comparison in this area.}
\end{itemize}

\paragraph{Efficient software engineering optimization (37.8\%)}
\begin{itemize}
\setlength{\itemsep}{0pt}
\setlength{\parskip}{0pt}
\setlength{\topsep}{0pt}
\item \textit{While our frameworks demonstrate strong performance, their computational overhead could be further optimized for efficiency.}
\item \textit{Our implementation was not optimized thoroughly and could be made significantly faster, which we defer to future research.}
\end{itemize}

\paragraph{Adversarial attack defense (37.3\%)}
\begin{itemize}
\setlength{\itemsep}{0pt}
\setlength{\parskip}{0pt}
\setlength{\topsep}{0pt}
\item \textit{Future work should evaluate the system under stronger adversarial settings.}
\item \textit{A systematic analysis is needed to better understand offensive or unsafe content in the dataset.}
\end{itemize}

%----------------------------------------------------------------------------------%
\section{Author Rankings}
\label{app:author-rankings}

\begin{table}[t]
\centering
\small
\begin{tabular}{lrrr}
\toprule
\textbf{Author} & \textbf{Papers} & \textbf{FW items} & \textbf{Addressed} \\
\midrule
Zhiyuan Liu    & 234 & 459 & 283 \\
Maosong Sun    & 250 & 449 & 243 \\
Dan Roth       & 315 & 375 & 241 \\
Graham Neubig  & 301 & 398 & 239 \\
Noah A. Smith  & 252 & 338 & 235 \\
Iryna Gurevych & 384 & 476 & 230 \\
Luke Zettlemoyer & 164 & 321 & 219 \\
Mirella Lapata & 204 & 301 & 212 \\
\bottomrule
\end{tabular}
\caption{Authors with the most future work items later addressed by the community.}
\label{tab:author_proposers}
\end{table}

\begin{table}[t]
\centering
\small
\begin{tabular}{lrr}
\toprule
\textbf{Author} & \textbf{Papers} & \textbf{Others' FW addressed} \\
\midrule
Dan Roth       & 315 & 1,206 \\
Reut Tsarfaty  &  87 & 1,125 \\
Mor Geva       &  47 & 1,093 \\
Yoav Goldberg  & 163 & 1,091 \\
Ashish Sabharwal & 45 & 1,070 \\
Tushar Khot    &  35 & 1,057 \\
Tomer Wolfson  &   8 & 1,042 \\
Harsh Trivedi  &  13 & 1,031 \\
\bottomrule
\end{tabular}
\caption{Authors whose papers address the most FW items proposed by others.}
\label{tab:author_addressing_others}
\end{table}
We consider two complementary signals: how many of an author's own FW items are later addressed by the community, and how many FW items proposed by others are addressed by that author's papers. These measures capture different forms of participation in research agenda formation: proposing directions that others pursue, and contributing work that responds to directions proposed elsewhere. Table \ref{tab:author_proposers} reports authors with the most future work items later addressed by the community, while Table \ref{tab:author_addressing_others} reports authors whose papers address the most future work items proposed by others.

These rankings should not be read as measures of researcher quality. They are affected by productivity, co-authorship structure, topic popularity, field size, and the time span covered by the corpus. Their value is descriptive: they show that agenda formation is unevenly distributed, with some authors repeatedly proposing directions that others take up, and others publishing work that aligns with many previously stated directions.

\end{document}